\documentclass{article}

\usepackage{PRIMEarxiv}

\usepackage[utf8]{inputenc} 
\usepackage[T1]{fontenc}    
\usepackage{hyperref}       
\usepackage{url}            
\usepackage{booktabs}       
\usepackage{amsfonts}       
\usepackage{nicefrac}       
\usepackage{microtype}      
\usepackage{lipsum}
\usepackage{fancyhdr}       
\usepackage{graphicx}       
\graphicspath{{media/}}     

\usepackage{amsmath,amssymb,amsfonts}
\usepackage{algorithmic}
\usepackage{textcomp}
\usepackage{xcolor}
\usepackage[numbers]{natbib}

\usepackage{combelow}

\title{A Distributed Automatic Domain-Specific Multi-Word Term Recognition Architecture using Spark Ecosystem}

\author{
  Ciprian-Octavian Truic\u{a}$^{a}$, Neculai-Ovidiu Istrate$^{b}$, and Elena-Simona Apostol$^c$\\
  University Politehnica of Bucharest, Bucharest, Romania \\
  \texttt{$^a$ciprian.truica@upb.ro, $^{b}$neculai.istrate@stud.acs.upb.ro, $^c$elena.apostol@upb.ro}
}

\begin{document}
\maketitle

\begin{abstract}
Automatic Term Recognition is used to extract domain-specific terms that belong to a given domain.
In order to be accurate, these corpus and language-dependent methods require large volumes of textual data that need to be processed to extract candidate terms that are afterward scored according to a given metric.
To improve text preprocessing and candidate terms extraction and scoring, we propose a distributed Spark-based architecture to automatically extract domain-specific terms.
The main contributions are as follows: 
(1) propose a novel distributed automatic domain-specific multi-word term recognition architecture built on top of the Spark ecosystem;
(2) perform an in-depth analysis of our architecture in terms of accuracy and scalability;
(3) design an easy-to-integrate Python implementation that enables the use of Big Data processing in fields such as Computational Linguistics and Natural Language Processing.
We prove empirically the feasibility of our architecture by performing experiments on two real-world datasets.
\end{abstract}

\keywords{
    Automatic Term Recognition \and
    Spark Ecosystem \and
    Distributed System
}

\maketitle

\section{Introduction}\label{sec:introduction}

Automatic Term Recognition (ATR) is used to extract domain-specific terms that create the terminology of the domain.
A term can be defined as a linguistic structure or a concept and it is composed of one or more words with a specific meaning to a domain.
With the exponential growth of technical and scientific articles, new domain-specific terms appear daily as 
named entities (e.g., Apache Spark), idioms (e.g., Big Data), multi-word expressions (e.g., recurrent neural networks), or through semantic change and shifts (e.g., local neighborhood).
Methods that can automatically recognize and extract these domain-specific terms are useful for both scientists and professionals to improve existing systems (i.e., WordNet~\cite{Chiru2021}, OntoLex-FRaC~\cite{Chiarcos2022}) that deal with linguistics, terminology, and machine-readable technologies.

ATR methods~\cite{Frantzi1997,Frantzi1999,Frantzi2000,LossioVentura2013,LossioVentura2014,LossioVentura2015} consist of two main phases.
The first phase is extracting a list of candidate terms that will later be used by scoring metrics to rank their importance to a given domain.
To extract this list, words are tagged with their part of speech (PoS), and candidate multi-word terms are extracted using language-dependent linguistic filters~\cite{LossioVentura2014b}. 
The second phase is specific to each method and involves computing a score of domain relevance by using different term statistics, e.g., frequency, context, number of similar terms, etc.

Users can process large volumes of textual data when employing ATR methods.
The extraction and recognition of domain-specific terms can be improved by developing the methods on top of distributed ecosystems such as Apache Hadoop and Apache Spark.
Thus, in this work, we try to address the following research question :
\begin{itemize}
    \item[($Q_1$)] Can we develop an optimized distributed domain-specific muti-word terms recognition architecture?
    \item[($Q_2$)] Can we improve the time performance of automatic term extraction using a distributed environment?
\end{itemize}

To answer the two research questions, we propose the use of distributed systems for the fast computation of scores for ranking domain-specific multi-word terms.
Furthermore, we aim to improve data availability and create a fault-tolerant system that uses data duplication by employing a distributed database management system to store the results of the ATR methods.
Furthermore, we aim to test our architecture to prove its feasibility in real-world scenarios.
Thus, the main objectives of this work are as follows:
\begin{itemize}
    \item[($O_1$)] Develop a distributed automatic domain-specific multi-word term recognition architecture using a distributed environment;
    \item[($O_2$)] Perform extensive experiments to prove empirically the feasibility of the proposed architecture.
\end{itemize}

By developing an ATR system on top of the Hadoop Ecosystems, employing Apache Spark framework for processing and MongoDB NoSQL database for storing, we try to achieve the two main objectives of this work.
We use two real-world datasets to test empirically the accuracy and scalability of the proposed system.
We summarize the main contributions of this work as follows:
\begin{itemize}
    \item[($C_1$)] Design a distributed text preprocessing and automatic domain-specific multi-word term recognition architecture deployed in an Apache Spark ecosystem and using a MongoDB database cluster;
    \item[($C_2$)] Present an in-depth benchmark using real-world datasets to showcase the accuracy and scalability of our architecture;
    \item[($C_3$)] Provide a Python implementation that uses the capabilities of Big Data processing platforms that can be easily integrated into other Natural Language Processing, Computational Linguistics, and Data Science tasks.
\end{itemize}

This paper is structured as follows.
Section~\ref{sec:stateoftheart} presents the current state-of-the-art methods for ATR.
Section~\ref{sec:metodology} describes the proposed architecture and its implementation on an Apache Hadoop cluster using Python, Apache Spark framework, and a MongoDB cluster.
Section~\ref{sec:results} presents the two real-world datasets used for experiments and discusses the results.
Section~\ref{sec:conclusions} concludes this work and hints at future work.

\section{State of the Art}~\label{sec:stateoftheart}

In the current literature, multiple Automatic Term Recognition methods have been proposed to extract domain-specific muti-word terms~\cite{Frantzi1997,Frantzi1999,Frantzi2000,LossioVentura2013,LossioVentura2014,LossioVentura2015}.
These methods rely on word statistics and linguistic filters that are corpus and language dependent and use Automatic Term Extraction (ATE) methods for scoring candidate terms~\cite{Manning2008,SparckJones2000a,SparckJones2000b,Paltoglou2010}, e.g., frequency, term frequency (TF), inverse document frequency (IDF), term frequency-inverse document frequency (TFIDF), OkapiBM35, etc.
Most of these methods were developed for extracting and improving the linguistic taxonomies and vocabularies with the terminology used in the bio-medical domains.
As they are corpus dependent and rely solely on statistical measures of words within the set of texts used, they generalize well, and they can be applied to other domains.
ATR systems are sensitive to variations in language and require a large amount of data to manage to extract terms effectively.
Also, ATR methods are not trained, they rely entirely on the corpus statistic. Thus, when presented with new terms these methods cannot determine if they are domain-specific or not.

C-Value~\cite{Frantzi1997} is one of the first methods proposed in the literature to extract domain-specific terms.
This method relies on linguistic filters to extract candidate terms.
Each term is afterward scored using the C-Value metric that employs candidate term frequency and nested term frequency.

An extension to this method is NC-Value~\cite{Frantzi1999,Frantzi2000}, which introduces the concept of context terms.
Context terms are terms that have a higher impact than others on the studied domain.
These terms are either given by an expert or are inferred using the C-Value.
After constructing a list of context terms, they are weighted using their co-occurrence in the set of documents. 
The NC-Value scores candidate terms based on the C-Value score of candidate terms and the weight of the context terms that appear in the candidate terms.

C-TFIDF and F-Okapi~\cite{LossioVentura2013} are two methods that use the TFIDF and OkapiBM25 computed for each candidate term instead of frequency to compute the C-Value.
These two ATE metrics are employed to minimize the bias introduces by common words and by the length of documents.

F-TFIDF and F-Okapi~\cite{LossioVentura2013} are two mode methods that combine C-Value with the ATE methods TFIDF and OkapiBM35.
The candidate terms are extracted as usual using linguistic patterns and scored with C-Value.
Both metrics are defined as a harmonic mean between the C-Value and TFIDF, respectively OkapiBM25.

LIDF-Value~\cite{LossioVentura2014} is another ATR method that incorporates the importance of a linguistic feature to extract candidate terms and minimizes the bias introduced by common words.
The metric uses IDF to increase the importance of candidate terms and also computes a probability for each linguistic pattern.

TerGraph~\cite{LossioVentura2014} is a graph-based ATR method that tries to deal with the following issues that the other methods have: 
(1) extraction of non-valid terms (noise) 
(2) omission of terms with low frequency (silence),
(3) extraction of multi-word terms having complex and various structures, and
(4) validation efforts of the candidate term.
The method relies on LIDF-Value to extract a list of candidate terms and then constructs a graph where nodes are the terms and the vertices are their co-occurrence in the corpus sentences.
These values are then used to compute the TerGraph metric of a candidate term based on its neighbors.

\section{Architecture}~\label{sec:metodology} 

\begin{figure}[!ht]
    \centering
    \includegraphics[width=0.75\columnwidth]{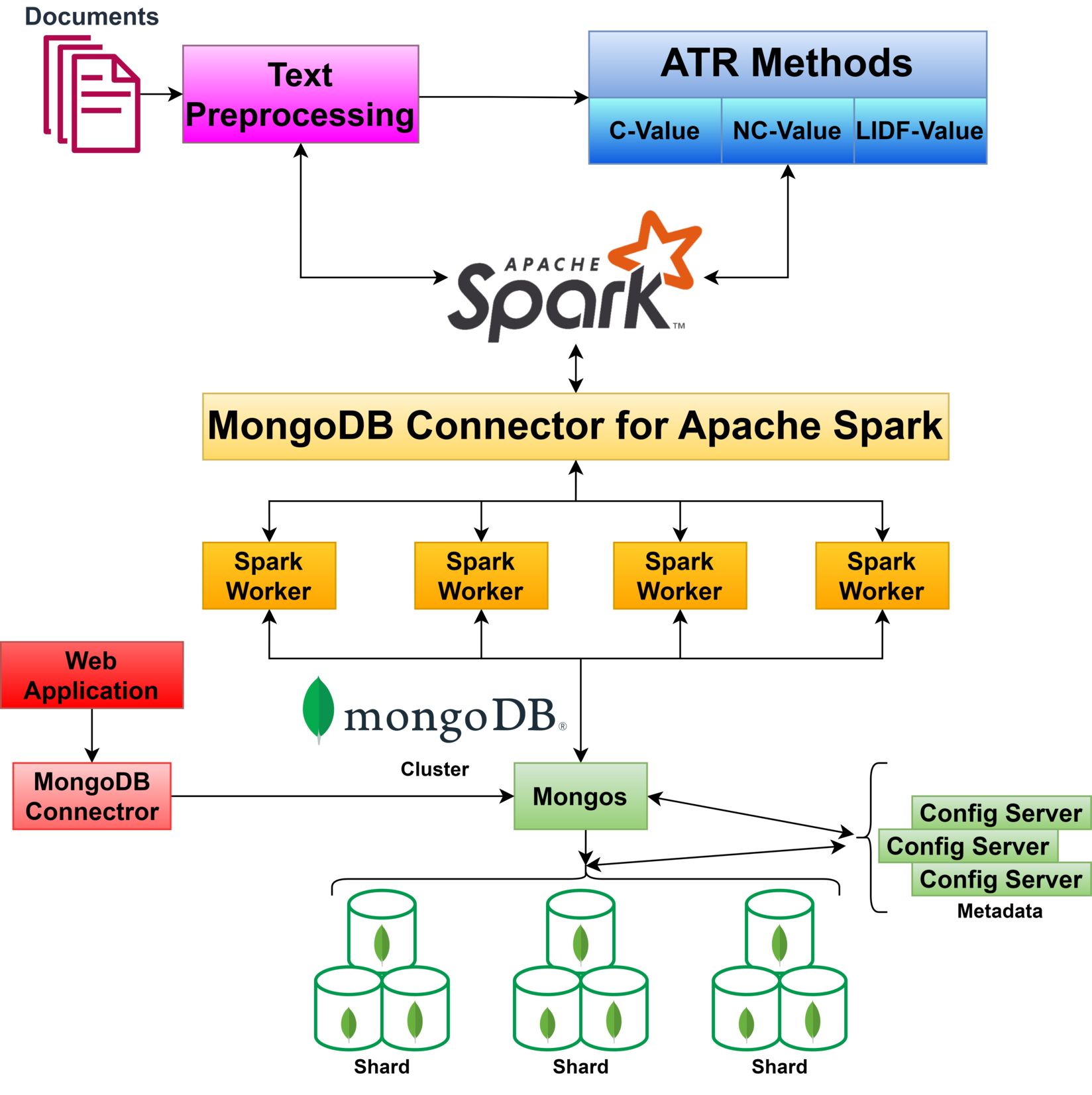}
    \caption{The Architecture}
    \label{fig:Architecture}
\end{figure}

Figure~\ref{fig:Architecture} presents the proposed architecture for automatic term recognition.
A corpus of documents is uploaded in the Text Preprocessing Module.
The preprocessing includes a pipeline implemented in Apache Spark for text cleaning that extracts the candidate terms and their frequencies and stores them in a distributed MongoDB database.
The Automatic Term Recognition (ATR) Module queries the database and computes for each candidate term different multi-word expression metrics using the Apache Spark Environment and then updates the records in the database.
We also provide a Web Application with a Graphical User Interface to query the results stored in the distributed MongoDB cluster for easy user interaction and analysis.

\subsection{Text Preprocessing module} 

This module uses Apache Spark to preprocess the text and extract a list of candidate terms.
The preprocessing steps are as follows:
(1) replace special characters and digits with spaces;
(2) convert the text to lowercase, as the Automatic Term Recognition methods are not case-sensitive;
(3) remove stopwords;
(4) extract the part of speech (PoS) for each word;
(5) lemmatized the text to remove language inflections; and
(6) apply linguistic filters to extract candidate terms.

We use five linguistic filters $\varphi \in \Phi$, where $\Phi$ is the set of filters, to extract the candidate terms:
\begin{enumerate}
    \item[(1)] Noun followed by at least another noun (can be followed by any number of nouns);
    \item[(2)] At least one adjective followed by at least one noun;
    \item[(3)] One noun followed by at least one adjective followed by any number of nouns and adjectives pairs (in that order);
    \item[(4)] At least one noun followed by at least one adverb followed by any number of nouns; and
    \item[(5)] One noun followed by at least one verb followed by any number of adverbs and any number of adjectives.
\end{enumerate}

The filters use regular expressions to match the PoS for each lemma and extract the candidate terms.
Any number of filters can be defined using this syntax to address application-specific needs, but we limit this work to only the five filters presented above.
During this step, we also compute for each candidate term the frequency in the whole corpus $f(a)$, the number of documents they appeared in $n_a$, the inverse document frequency (IDF), and the number of the linguistic filter used to obtain that term.

Using the extracted information, we create a JSON document for each candidate term.
These documents are then stored in the distributed document-oriented database MongoDB as a collection of documents.
When the preprocessing of the collection is finished, the database contains a complete list of candidate terms that are ready to be processed by the ATR module.

\subsection{Automatic Term Recognition module} 

This module uses ATR methods to score the candidate terms and extract the domain-specific multi-word terms.
We employ the following metrics:
(1) C-Value;
(2) NC-Value; and
(3) LIDF-Value.
To process these metrics, we query the database to obtain the results of the Text Preprocessing module. 
We compute each metric using the Apache Spark framework. 
The ATR scores for each multi-word term are stored in the MongoDB database by updating the corresponding record.

\subsubsection{C-Value} 
C-Value~\cite{Frantzi2000} is an automatic method used to extract domain-specific multi-word terms focusing on the extraction of nested terms, i.e., terms that can be found inside other terms.
The method relies on linguistic patterns to extract multi-word term candidates and their frequency.
As it is a statistical method, C-Value is domain agnostic which means that it can be applied to different fields with good results.
However, to have good results, all the corpus documents must belong to the same domain. 
If a corpus contains documents from different domains, the frequency of individual domain-specific terms is diminished, thus, affecting the accuracy of the C-Value score.
Equation~\eqref{eq:cvalue} presents the mathematical formula for C-Value for extracting both single and multi-word domain-specific terms (i.e., $a$) as proposed in~\cite{Truica2021b}, where: 
\begin{itemize}
    \item $|a|$ is the number of words that this candidate term has;
    \item $f(a)$ is the frequency with which the term has appeared in the corpus;
    \item $T_{a}$ the list of terms in which the term $a$ appears as a nested term; 
    \item $P(T_{a})$ the cardinality of the $T_{a}$. 
\end{itemize}

\begin{equation}\label{eq:cvalue}
    \text{C-Value}(a) = \begin{cases}
        log_{2}|1 + a| \cdot f(a), \text{if } f(a) \text{ is not nested}\\
        log_{2}|1 + a| \cdot \left ( f(a) - \sum_{b \in T_a} \dfrac {f(b)} {P(T_a)} \right ), \text{otherwise}
    \end{cases}
\end{equation}

\subsubsection{NC-Value} 

NC-Value \cite{Frantzi2000} is an Automatic Term Recognition method that uses C-Value and incorporates the contextual information of the words, as words found in similar contexts tend to be more relevant to domain-specific terms.
The first step of computing NC-Value is to compute C-Value and extract a list of candidate terms. 
The second step involves the construction of a list of context words $C$.
The list contains the individual words that belong to the top (or all) candidate terms extracted and scored with C-Value.
Each context word is them weighted using Equation~\eqref{eq:weight}, where
\begin{itemize}
    \item $c \in C$ is the context word;
    \item $n(c)$ is the number of terms in which the word $c$ appears; 
    \item $n_C$ is the number of considered context terms.
\end{itemize}

\begin{equation}\label{eq:weight}
    w(c) = \frac{t(c)}{n_C}
\end{equation}

The final step involves the actual calculation of the NC-Value score for the candidate multi-word term $a$ using the formula in Equation~\eqref{eq:ncvalue}, where \begin{itemize}
    \item $C_a$ is the list of distinct contextual words of $a$;
    \item $b$ is a context word in $C_a$;
    \item $f_a(b)$ is the frequency of $b$ as a term context word of $a$.
\end{itemize}

\begin{equation}\label{eq:ncvalue}
    \text{NC-Value}(a) = 0.8 \cdot \text{C-Value}(a) + 0.2 \cdot \sum_{b \in C_a} {{f_a}(b)} \cdot w(b)
\end{equation}

The hyper-parameter values of 0.2 and 0.8 are chosen empirically to get higher accuracy.
It is important to note that NC-Value only applies different weights to some of the candidate terms already extracted by C-Value.
Thus, no new terms are extracted but the final list of domain-specific terms might be reordered using the NC-Value score.

\subsubsection{LIDF-Value}

LIDF-Value~\cite{LossioVentura2014} is another C-Value based method that incorporates the relevance of a term using 
(1) the probability $P(\varphi(a))$ of the linguistic filter $\varphi \in F$ used to determine the term $a$ and 
(2) the importance of the term in a set of given documents using the inverse document frequency ($IDF(a)$). 
$P(a_{\varphi})$ is computed as the frequency $f(\varphi(a))$ of the linguistic filter $\varphi \in \Phi$ used to determine the multi-word term $a$ over the sum of all linguistic filter frequencies $\Sigma_{\varphi' \in {\Phi}} f(\varphi')$ (Equation~\eqref{eq:plfa}).
This probability penalizes the terms that appear as ``noise", i.e., terms that do not belong to the domain but have the same linguistic filter as those that are.

\begin{equation}\label{eq:plfa}
    P(\varphi(a)) = \frac{f(\varphi(a))}{\Sigma_{\varphi' \in {\Phi}} f(\varphi')}
\end{equation}

$IDF$ measures how common or rare a word or a multi-word term is w.r.t. the entire corpus documents.
Equation~\eqref{eq:idf} presents the formula for computing $IDF(a)$ for a multi-word term $a$, where
\begin{itemize}
    \item $N$ is the number of documents in the corpus;
    \item $n(a)$ is the number of documents containing $a$.
\end{itemize}

\begin{equation}\label{eq:idf}
    IDF(a) = \log\left(\frac{N} {n(a)}\right)
\end{equation}

Using the $P(\varphi(a))$ and the $IDF(a)$, we can compute the LIDF-value for a multi-word term $a$, as shown in  Equation~\eqref{eq:lidfvalue}:

\begin{equation}\label{eq:lidfvalue}
    \text{LIDF-value}(a) = {{P(\varphi(a))} \cdot IDF(a) \cdot \text{C-Value}(a)}
\end{equation}

\subsection{Spark Environment}

We use a distributed Apache Spark~\cite{zaharia2010spark} for the text preprocessing pipeline and to compute the ATR scores for the candidate terms.
The Spark cluster contains six nodes: one node is used as the Namenode, one node is used as the Secondary Namenode, and all the nodes are used ad Datanodes.
We use YARN~\cite{Vavilapalli2013} as the resource manager.
On the cluster, we also have installed HDFS with a replication factor of three.
The datasets are residing on the HDFS prior to upload and text preprocessing.

\subsection{Database} 

\begin{figure}[!htbp]
    \centering
    \includegraphics[width=0.5\columnwidth]{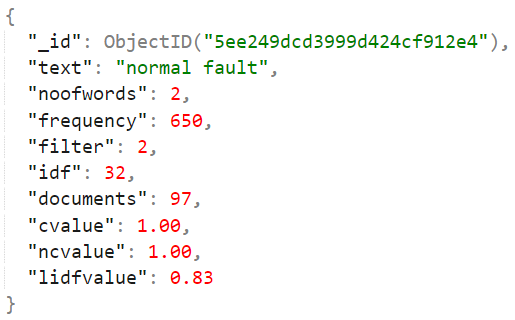}
    \caption{MongoDB document}
    \label{fig:mongodoc}
\end{figure}

The results of the Text Preprocessing and ATR modules are stored in a distributed MongoDB database cluster. 
MongoDB is a NoSQL database that stores the data by using a Binary JSON, i.e., BSON, document structure~\cite{Truica2013,Truica2015}.
We use a distributed cluster to offer fault-tolerance, high availability, and data redundancy~\cite{Truica2020,Truica2021}.
The information stored in the database resides in a sharded collection that contains BSON documents with all the information for a candidate multi-word term (Figure~\ref{fig:mongodoc}).

The MongoDB cluster is composed of three shards (Figure~\ref{fig:Architecture}), one \textit{mongos} instance, and a Configuration Servers deployed as a Replica Set.
Each shard is built on top of a Replica Set with one primary node and two secondary nodes to offer data redundancy and fault tolerance within the shard.
The \textit{mongos} instance acts as the interface between the client application and the sharded cluster, while the Configuration Server Replica Set stores the metadata for a sharded cluster.
The connection between the Text Preprocessing and ATR modules is done using the MongoDB Connector for Apache Spark.

\subsection{Implementation}

The architecture is implemented using Python.
We chose this programming language as it is wildly used in the field of Machine Learning, Deep Learning, Computational Linguistics, and Natural Language Processing.
Our implementation can be easily integrated into other systems by scientists and practitioners in these fields.
Thus, we enable the use of Big Data processing platforms in these fields.

A dataset residing on HDFS or on the local HDD is uploaded in the Text Preprocessing module using the command line interface.
To process the dataset using the Apache Spark environment, we employ \href{https://spark.apache.org/docs/latest/api/python/}{PySpark}.
Both modules use their own Spark Session to distribute the text preprocessing pipeline and the automatic term recognition methods, respectively.
To read and store data, the Spark Session uses MongoDB Connector for Apache Spark.
The data is read as a Spark DataFrame~\cite{Armbrust2015}.
For preprocessing and computing the automatic term recognition metrics, we transform the DataFrame into an RDD~\cite{Zaharia2012} object to use native transformation and action functions.

The Text Preprocessing module starts a Spark Session to apply the different steps of the preprocessing pipeline.
This pipeline is implemented using \href{https://spacy.io/}{spaCy} and \href{https://www.nltk.org/}{NLTK}~\cite{Bird2009}.
The six preprocessing steps are encapsulated into a User Defined Function which is applied to the raw texts to obtain the candidate terms' information as follows: (1) the multi-word term, (2) the filter used to obtain the multi-word term, (3) the length of the multi-word term, (4) the multi-word term frequency, (5) the multi-word term IDF score, and (6) the number of documents where the multiword appears.
This information is saved as a collection in MongoDB.  
At the end of the preprocessing task, we close the Spark Session.

After preprocessing, we compute the different ATR scores.
We start a Spark Session to connect the MongoDB database and extract into a DataFrame the information previously stored by the Text Preprocessing module.
We transform the DataFrame into an RDD and we compute the scores for each row in the following order: C-Value, NC-Value, and LIDF-Value.
When the computation is done, we transform the RDD back into a DataFrame and update the existing MongoDB collection with the ATR scores.
At the end of the ATR task, we close the Spark Session.

\begin{figure}[!htb]
    \centering
    \includegraphics[width=0.75\columnwidth]{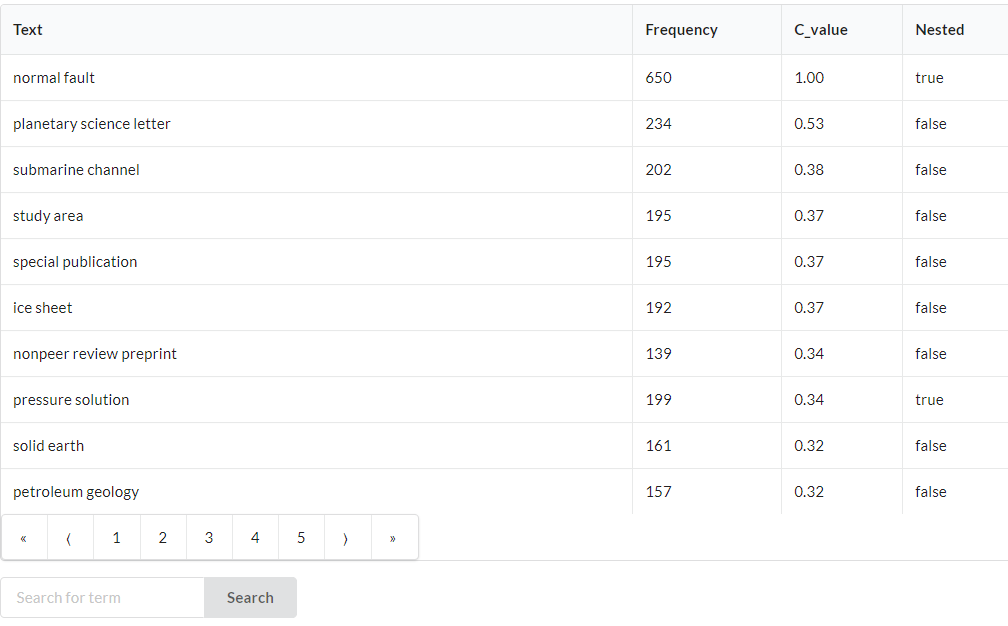}
    \caption{C-Value results from the client side}
    \label{fig:CValueREs}
\end{figure}

The Web Application is used to read the information directly from MongoDB and present it using a Graphical User Interface (GUI) implemented in \href{https://nodejs.org/en}{Node.js}.
The GUI provides users with the following functionalities: 
(1) visualize results for each score in different tabs (Figure~\ref{fig:CValueREs}); and
(2) search for a specific multi-word term.

\section{Experimental Results}~\label{sec:results}

This section presents the experimental results obtained with ATR.
For extracting and analyzing the ATR scores we use two datasets, i.e., Medical and Geology.
We also perform a scalability test to determine the time performance of our method when varying the number of documents in a dataset.

\subsection{Experimental conditions}

For the experiments, we use a cluster with 6 nodes running Ubuntu 22.04 x64.
The system uses a homogeneous architecture, thus each node has 1 Intel Core i7-4790S CPU with 8 cores at 3.20GHz, 16 GB RAM, and 500GB HDD.
We use Apache Ambari as a cluster manager.
The cluster has the following configuration:
(1) one node is Name Node, 
(2) one is the Secondary Name, and
(3) all the nodes act as HDFS Data Nodes and YARN Node Managers.
On the Name Node machine, we also started the YARN Resource Manager, YARN Application Manager, and Spark Driver.
YARN, Spark, and MapReduce Clients are installed on all the nodes.

The main function of the HDFS~\cite{Shvachko2010} Name Node is to keep all the information regarding the files in the distributed system.
Thus, this master node stores the data and metadata (e.g., names, locations, access dates, etc.) about the directory tree structure and files.
The HDFS Secondary Name Node is a master node that reduces the load of the Name Node by performing periodic namespace metadata checkpoints.
The worker nodes, i.e., HDFS Data Nodes, store and manage HDFS blocks on the local machine HDD.

The YARN Resource Manager~\cite{Vavilapalli2013} is a master node that manages the execution of applications on the cluster.
Each application is composed of a job that has multiple tasks to complete.
YARN Resource Manager schedules the jobs on the cluster, assigns resources to tasks on Data Nodes using the YARN Node Manager, and monitors their status.
The YARN Node Manager is a worker node that has two functions: (1) runs and manages tasks on an individual node, and (2) reports the health and status of the tasks in execution to the YARN Resource Manager.

We configure Spark's number of executors to be fixed to 16.
Each executor is configured to claim one vnode and 3GB of memory.
The Spark Driver, the master node for the application, is running on the same machine as the Name Node and uses the Task Scheduler to launch tasks via cluster manager, i.e., YARN.
Spark uses a Directed Acyclic Graph (DAG) Scheduler to divide application jobs into stages of tasks that are executed on the worker nodes, i.e., Spark Clients.

For the MongoDB cluster, we used a cluster infrastructure that contains 4 nodes with the same hardware configuration as the Hadoop cluster.
We configure one node to be the entry point to the cluster, i.e., \textit{mongos}.
On the same node, we also configure the primary replica node for the Configuration Server Replica set, while two secondary replicas are configured on 2 of the remaining three nodes.
The three Sharded Replica Sets are installed on the remaining nodes as follows:
(1) the primary replica for each shard resides on its own node;
(2) tow secondary nodes reside on the other two nodes that do not store the primary for the shard that they belong.
Thus, each machine is configured to have an instance of a primary replica for a shard and two replicas, one for each of the other two shards.

\subsection{Automatic Term Recognition results}

This section presents the results obtained by each ATR metric.
The results are normalized by dividing all the scores obtained by a metric with the highest score obtained using that metric.

\subsubsection{Medical Dataset results}

The Medical dataset is collected from MedPud and contains approximately 5\,000 abstracts from which we extracted 637 candidate terms.

Table~\ref{tab:CValue_med} presents the top-10 multi-word terms from the medical domain extracted with C-Value.
We can observe that all the extracted terms, with the exception of ``numerical data" which is a term more related to the statistics domain, are related to the medical domain.
If a term is not nested, then the C-Value score decreases, e.g., ``radiation effects" has a lower score than the other nested terms that are ranked higher on the list.

\begin{table} [!htbp]
    \centering
    \caption{C-Value Top-10 Multi-Word Terms from Medical Dataset}\label{tab:CValue_med}
    \begin{tabular}{lccc}
        \hline
        \textbf{Multi-word term} & \textbf{C-Value} &  \textbf{Frequency} & \textbf{Nested Term?} \\
        \hline 
        drug effects	        & 1.00	& 8\,845	& Yes \\
        therapeutic use	        & 0.32	& 2\,874	& Yes \\
        drug therapy	        & 0.23	& 2\,067	& Yes \\
        binding proteins        & 0.15	& 1\,354	& Yes \\
        adverse effects	        & 0.15	& 1\,321	& Yes \\
        numerical data	        & 0.13	& 1\,139	& Yes \\
        radiation effects	    & 0.08	& 716	    & No  \\
        diagnostic imaging	    & 0.08	& 684	    & Yes \\
        dendritic cells	        & 0.05	& 398	    & No  \\
        plasmodium falciparum	& 0.04	& 316	    & No  \\
        \hline
    \end{tabular}
\end{table}

NC-Value obtains the same normalized results and top-10 multi-word terms as C-Value (Table~\ref{tab:NCValue_med}).
For this dataset, we observe that there is no ranking list reordering.
This result is influenced by the context terms and their weights.
In the case of this dataset, we used the list of terms extracted by C-Value to create the context term list.
Furthermore, the impact value of the score and the ranking of the list is also influenced by the number of candidate terms extracted by C-Value that are then used as context terms when employing NC-Value.
Thus, with a small number of candidate terms, the automatically extracted list of context terms is small and, when computing the weights for the context terms, the difference among the weights is very small which in turn do not have a high impact on the NC-Value score.

\begin{table} [!htbp]
    \centering
    \caption{NC-Value Top-10 Multi-Word Terms from Medical Dataset}\label{tab:NCValue_med}
    \begin{tabular}{lccc}
        \hline
        \textbf{Multi-word term} & \textbf{NC-Value} & \textbf{C-Value} \\
        \hline 
        drug effects	        & 1.00	& 1.00 \\
        therapeutic use	        & 0.32	& 0.32 \\
        drug therapy	        & 0.23	& 0.23 \\
        binding proteins	    & 0.15	& 0.15 \\
        adverse effects	        & 0.15	& 0.15 \\
        numerical data	        & 0.13	& 0.13 \\
        radiation effects	    & 0.08	& 0.08 \\
        diagnostic imaging	    & 0.08	& 0.08 \\
        dendritic cells	        & 0.05	& 0.05 \\
        plasmodium falciparum	& 0.04	& 0.04 \\
        \hline
    \end{tabular}
\end{table} 

Table~\ref{tab:LIDF_med} presents the top-10 terms extracted with LIDF-Value.
We can observe that the ranked list is reordered when using this metric.
Both the probability of the linguistic filter and the IDF score influence the position of the domain-specific terms in the list.
We observe that ``drug therapy" changes its position as it appears in fewer documents than ``therapeutic use".
We also observe that new domain-specific terms appear in the top-10 list, e.g., ``blood supply", while others are removed, e.g., ``dendritic cells".

\begin{table}[!htbp]
    \centering
    \caption{LIDF-Value Top-10 Multi-Word Terms from Medical Dataset}
    \label{tab:LIDF_med}
    \begin{tabular}{lccc}
        \hline
        \textbf{The Extracted term} &\textbf{LIDF-Value} & \textbf{C-Value}\\
        \hline 
        drug effects	      & 1.00	& 1.00 \\
        drug therapy	      & 0.36	& 0.23 \\
        therapeutic use	      & 0.30	& 0.32 \\
        radiation effects	  & 0.17	& 0.08 \\
        adverse effects	      & 0.16	& 0.15 \\
        binding proteins	  & 0.16	& 0.15 \\
        numerical data	      & 0.15	& 0.13 \\
        diagnostic imaging	  & 0.11	& 0.08 \\
        plasmodium falciparum & 0.09	& 0.04 \\
        blood supply	      & 0.08	& 0.03 \\
        \hline
    \end{tabular}
\end{table}

\subsubsection{Geology Dataset results}

The Geology dataset is collected from EarthArXiv and contains approximately 1\,000 full articles from which we extracted 128\,316 candidate terms.

In table~\ref{tab:CValue}, the top-10 terms obtained with C-Value are presented.
We observe that most of these terms are from the geology domain. 
C-Value also finds some multi-word terms from the scientific publishing vocabulary which are also found in geology articles, i.e., ``study area", ``special publication", and ``nonpeer review preprint".
These terms are selected as domain-specific due to their frequency in the dataset.
For example, the term ``normal fault'' has a much higher frequency which results in a far greater C-Value score.
Even if the method is trying to diminish the frequency influence by using other statistical values, the high frequency of this term is too hard to overcome even if it is a nested term.
Nested terms also have an impact on the final C-Value score, e.g., ``nonpeer review preprint'', a scientific publishing domain term, is not nested obtaining a higher C-Value score, while ``pressure solution", a geology domain term, is nested and has a smaller  C-Value score as it appears in other candidate terms such as ``pressure solution creep".

\begin{table} [!htbp]
    \centering
    \caption{C-Value Top-10 Multi-Word Terms from Geology Dataset}\label{tab:CValue}
    \begin{tabular}{lccc}
        \hline
        \textbf{Multi-word term} & \textbf{C-Value} &  \textbf{Frequency} & \textbf{Nested Term?} \\
        \hline 
        normal fault             & 1.00 & 650 & Yes \\
        planetary science letter & 0.53 & 234 & No  \\
        submarine channel        & 0.38 & 202 & No  \\
        study area               & 0.37 & 195 & No  \\
        special publication      & 0.37 & 195 & No  \\
        ice sheet                & 0.37 & 192 & No  \\
        nonpeer review preprint  & 0.34 & 139 & No  \\
        pressure solution        & 0.34 & 199 & Yes \\
        solid earth              & 0.32 & 161 & No  \\
        petroleum geology        & 0.32 & 157 & No  \\
        \hline
    \end{tabular}
\end{table}

Table~\ref{tab:NCValue} presents the top-10 terms obtained with the NC-Value method compared with the C-Value score.
We observe that the ranked top-10 list for NC-Value is not the same as for C-Value, as NC-Value tries to add more context when scoring candidate terms by weighting context terms.
To create the list of context terms, we use the existing candidate terms ranked by C-Value.
The ranking can be improved by using expert input to create the list of context terms.

\begin{table} [!htbp]
    \centering
    \caption{NC-Value Top-10 Multi-Word Terms from Geology Dataset}\label{tab:NCValue}
    \begin{tabular}{lccc}
        \hline
        \textbf{Multi-word term} & \textbf{NC-Value} & \textbf{C-Value} \\
        \hline 
        normal fault             & 1.00 & 1.00 \\
        planetary science letter & 0.51 & 0.53 \\
        submarine channel        & 0.38 & 0.38 \\
        study area               & 0.36 & 0.37 \\
        special publication      & 0.36 & 0.37 \\
        ice sheet                & 0.35 & 0.37 \\
        nonpeer review preprint  & 0.35 & 0.34 \\
        pressure solution        & 0.33 & 0.34 \\
        fault system             & 0.32 & 0.30 \\
        solid earth              & 0.31 & 0.32 \\
        \hline
    \end{tabular}
\end{table} 

Table~\ref{tab:LIDF} presents the top-10 terms obtained with the LIDF-Value method compared with the C-Value scores.
We observe that new terms appeared in the top rank list compared to C-Value/NC-Value, e.g., ``lobate example'' is ranked the new top-ranked term. 
The results are influenced by both the IDF and the probability of the filter used ($P(\varphi(a))$).
The ``lobate example" term appears in only one document, making the IDF value of this term to be significantly higher than, for example, ``normal fault" which is a common term in the corpus of documents.
Another example of the impact of these two factors that LIDF-Value considers is the term ``submarine channel", which is ranked $3^{rd}$ by C-Value and is ranked $9^{th}$ by LIDF-Value.
As with the other methods, we observe that terms from the scientific publishing domain are also selected by this method.

\begin{table}[!htb]
    \centering
    \caption{LIDF-Value Top-10 Multi-Word Terms from Geology Dataset}
    \label{tab:LIDF}
    \begin{tabular}{lccc}
        \hline
        \textbf{The Extracted term} &\textbf{LIDF-Value} & \textbf{C-Value}\\
        \hline 
        lobate example                  & 1.00 & 0.29 \\
        pressure solution               & 0.97 & 0.34 \\
        ice sheet                       & 0.89 & 0.37 \\
        sand body                       & 0.87 & 0.28 \\
        nonpeer review preprint upload  & 0.84 & 0.19 \\
        salt wall                       & 0.83 & 0.29 \\
        normal fault                    & 0.83 & 1.00 \\
        nonpeer review preprint publish & 0.82 & 0.22 \\
        submarine channel               & 0.81 & 0.38 \\
        test case                       & 0.80 & 0.18 \\
        \hline
    \end{tabular}
\end{table}

\subsection{Scalability Results}

To correctly evaluate the time performance of the Text Preprocessing module and each ATR score, we create separate classes for each metric and store each metric in its own MongoDB collection.
The code for this set of experiments is available online on GitHub at \url{https://github.com/DS4AI-UPB/ATR-pySpark}.

In our experiments, the NC-Value and the LIDF-Value use the already computed C-Value.
The experiments are done on 4 datasets created by duplicating the Medical dataset to obtain different scalability factors. 
Table~\ref{tab:scalability} presents the time performance of text preprocessing and each individual score.

We observe that text preprocessing takes the longest, as is expected due to the large number of transformations required to obtain the candidate terms for each filter and their corpus metrics (i.e., frequency, IDF, and number of documents each candidate term appears) from the raw text.

When analyzing the time performance of the metrics, we observe that C-Value takes the longest to compute.
The computation performance of C-Value is increased in the case that a candidate term is also a nested term.
The decrease in time performance is almost linear with the size of the dataset.
For NC-Value and LIDF-Value, we observe that the metrics are almost constant.
The small differences among them are given by the time required for the Spark Workers to initialize.

\begin{table}[!htb]
    \centering
    \caption{Scalability results over 10 runs (time are in seconds)}
    \label{tab:scalability}
    \begin{tabular}{ccccc}
        \hline
\begin{tabular}[c]{@{}c@{}}\textbf{Number of}\\ \textbf{Documents}\end{tabular} 	& \begin{tabular}[c]{@{}c@{}}\textbf{Text}\\ \textbf{Preprocessing}\end{tabular} 	& \textbf{C-Value}			& \textbf{NC-Value}			& \textbf{LIDF-Value}	  \\ \hline
 5\,000		   & 34.57 $\pm$ 0.31	& 15.31 $\pm$ 1.02 	& 6.17 $\pm$ 0.13	& 6.93 $\pm$ 0.71 \\ 
10\,000			& 45.34 $\pm$ 0.21	& 16.55 $\pm$ 2.12 	& 6.20 $\pm$ 0.15	& 7.17 $\pm$ 0.65 \\ 
15\,000			& 56.46 $\pm$ 0.41	& 17.29 $\pm$ 0.90 	& 6.19 $\pm$ 0.17	& 6.91 $\pm$ 0.55 \\
20\,000			& 64.80 $\pm$ 0.69	& 18.81 $\pm$ 0.31 	& 6.86 $\pm$ 0.21	& 7.16 $\pm$ 0.27 \\ 	
25\,000			& 89.01 $\pm$ 0.71	& 19.50 $\pm$ 0.60	& 6.47 $\pm$ 0.41	& 6.89 $\pm$ 0.69 \\	
        \hline
    \end{tabular}
\end{table} 

\section{Conclusions}~\label{sec:conclusions}

In this paper, we present a novel automatic domain-specific multi-word expression recognition architecture implemented in Python and distributed using the Spark environment, thus answering ($Q_1$) and achieving objective ($O_1$).
We perform an in-depth analysis on two real-world datasets as well as scalability experiments to prove empirically the feasibility of the proposed architecture, thus answering ($Q_2$) and achieving objective ($O_2$).

The experimental results show that the proposed architecture manages to extract domain-specific accurately using the three ATR metrics.
We observe that a good strategy is to use the three different metrics together rather than separately, as when analyzing the combined results the overall accuracy of extracting domain-specific multi-word terms increases. 

We offer a Python implementation for our proposed architecture that can be easily used by scientists and practitioners for other Natural Language Processing, Computational Linguistics, and Data Science tasks.

In future work, we aim to implement an automatic way for acronym expansion to improve domain-specific term recognition. 
By expanding acronyms, we enhance the candidate terms and the context term lists, thus improving the accuracy of the ATR task.


\end{document}